\documentclass[10pt]{article}
\usepackage[preprint]{tmlr}

\usepackage{booktabs}
\usepackage{amsmath}
\usepackage{graphicx}
\usepackage{hyperref}
\usepackage{url}

\title{A Functional Preservation Score for Auditing Connectome Operations:\\
Compression, Alignment, Transport, and Local-Rule Writing}

\author{\name Praveg Vashishtha \email praveg\_pa2503mth259@iitp.ac.in \\
  \addr Department of Computer Science and Engineering, Indian Institute of Technology Patna, India \\
  \addr Sopra Banking Software India}

\def\month{11}
\def\year{2026}
\def\openreview{\url{https://openreview.net/forum?id=XXXXXXXX}}

\begin{document}
\maketitle

\begin{abstract}
We introduce the Functional Preservation Score (FPS), a null-normalised, behaviourally grounded
measure of how much of a neural network's input--output function survives a structural operation,
and use it to audit the operations underlying speculative ``mind-migration'' pipelines on public
data that runs on a single laptop. Across five pre-registered studies on the \emph{Caenorhabditis
elegans} connectome, whole-brain calcium activity, and hippocampal replay, FPS yields two positive
characterisations and a three-part negative. Low-bit quantisation of a real connectome exhibits a
structure-dependent sparsity floor: uniform int4/ternary quantisation is forced to roughly
$73\%/98\%$ effective sparsity because heavy-tailed integer synaptic weights round to zero, while
mixed-precision protection of hub neurons lowers these floors to about $27\%/32\%$. Decoded
hippocampal replay sequences compress heavily yet remain identifiable at a cross-validated top-1
accuracy of $0.69$ (95\% CI $[0.63,0.74]$; chance $0.125$; permutation $p<10^{-3}$) from latents of
order $10$--$50$ bits per event, and a bounded incremental store shows a graceful-versus-cliff
capacity trade-off. The migration pipeline fails in simulation at each stage for largely
independent reasons: the putative ``identity payload'' is not a functionally separable module
($p\geq0.16$ at every tested size); cross-individual functional transport via Gromov--Wasserstein
collapses once the target's wiring is respected (fidelity $1.00\!\to\!0.40$, $[0.31,0.49]$,
$p<10^{-3}$) or alignment is imperfect ($0.06$, $[0.00,0.13]$); and biologically plausible local
plasticity cannot reach the function the target topology provably admits (gap $0.45$,
$[0.27,0.64]$, $p<10^{-3}$, $d=2.0$). We release FPS, the data loaders, and all operations as a
toolkit. All results are at nematode / small-network scale and concern coherence in simulation,
not the feasibility of reading or writing a living brain.
\end{abstract}

\section{Introduction}
Structural operations on neural networks --- pruning, quantisation, alignment across instances,
and activity-driven weight installation --- appear across model compression, connectomics, and
speculative proposals to ``migrate'' a mind from one substrate to another. Such proposals combine
connectomics, graph neural networks, memory prosthetics, neural interfaces, and optogenetics into
a single partition--align--write pipeline, typically relying on the most favourable reading of
each component. Evaluating these operations requires a common question: how much of a network's
\emph{function} does an operation preserve? We make that question measurable with the Functional
Preservation Score (FPS) and apply it, on public data that runs on a laptop, both to operations
that are genuinely useful (compression) and to the stages of the migration pipeline that lack
empirical support.

Our contributions are: (1) FPS, a substrate-agnostic functional-preservation measure with an
explicit trivial-baseline null (\S\ref{sec:fps}); (2) a characterisation of connectome
compression, including a quantisation sparsity floor (\S\ref{sec:t1}); (3) a characterisation of
replay-sequence compressibility and a capacity--interference relationship (\S\ref{sec:t3}); (4) a
three-stage in-silico test of the partition--align--write pipeline that localises the failure mode
at each stage (\S\ref{sec:stages}); and (5) a reproducible toolkit (\S\ref{sec:repro}). The
unifying technical object is FPS used as a common currency across otherwise disparate operations.

\section{Related work}
Generic graph-network pruning and quantisation \citep{gnnprunequant2025} and scalable
Gromov--Wasserstein alignment \citep{xu2019gwl,xu2019sgwl,peyre2016gw} are established. Brain
alignment via optimal transport has been demonstrated at the scale of functional parcels
\citep{thual2022fugw}. Connectome-driven models can reproduce behaviour once dynamical parameters
are fitted \citep{dorkenwald2024flywire,flygnn2025}. The lottery-ticket literature
\citep{frankle2019lottery} studies sparse trainable subnetworks. Our contribution is not a new
compression or transport algorithm but a functional evaluation of these operations on connectome
substrates, and an explicit test of migration stages that are frequently asserted but, to our
knowledge, never measured. We summarise the biological claim provenance (synaptic turnover
\citep{vonderohe2006torpor}; memory prosthetics \citep{hampson2018prosthetic}; surface neural
interfaces \citep{jung2025bisc}) in Appendix~\ref{app:prov}.

\section{The Functional Preservation Score}\label{sec:fps}
Let a system map a fixed battery of stimuli to outputs. For a reference system with outputs
$Y_{\mathrm{ref}}$ and a transformed system with outputs $Y$, let
$\mathrm{raw}(Y)=1-\lVert Y-Y_{\mathrm{ref}}\rVert^2/\lVert Y_{\mathrm{ref}}-\overline{Y_{\mathrm{ref}}}\rVert^2$.
FPS normalises this by the best trivial baseline $Y_{\mathrm{null}}$, the larger of a shuffled
stimulus--response mapping and a predict-the-mean baseline:
\begin{equation}
\mathrm{FPS} = \mathrm{clip}\!\left(\frac{\mathrm{raw}(Y)-\mathrm{raw}(Y_{\mathrm{null}})}{1-\mathrm{raw}(Y_{\mathrm{null}})},\,0,\,1\right).
\end{equation}
An FPS of $1$ reproduces the reference function; $0$ is no better than the best trivial control.
Normalising by the null prevents an input-independent but plausible-looking output from scoring
well. For identity-style tasks we use identification accuracy (is reconstruction $i$ nearest its
own original among all events?), which is bounded with exact chance $1/n$.

All headline numbers are reported as a point estimate with a percentile bootstrap 95\% CI
($10^4$ resamples) over the relevant randomisation; comparisons use one-sided permutation tests
($10^4$ permutations, add-one-corrected) and Cohen's $d$ where applicable. To avoid selection
bias we report a pre-registered primary policy rather than a maximum over variants, and the
identification frontier uses 30-split repeated cross-validation with models fit on the training
split only.

\section{Study 1: Connectome compression}\label{sec:t1}
\begin{figure}[t]
\centering
\includegraphics[width=0.5\linewidth]{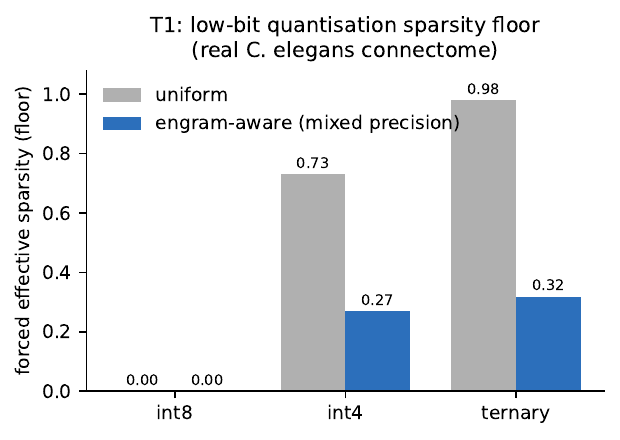}
\caption{Effective-sparsity floor by bit depth on the real connectome: uniform quantisation
(grey) versus engram-aware mixed precision (blue).}
\label{fig:t1}
\end{figure}

We load the real \emph{C. elegans} somatic connectome
\citep{white1986structure,cook2019connectome} (306 neurons; 81 sensory, 95 interneuron, 130 motor)
as a leaky-rate recurrent model and score the motor readout under compression. At matched effective
sparsity (int4), the pre-registered engram-aware policy preserves function better than uniform
pruning: FPS $0.30$ ($[0.12,0.48]$) versus $0.03$ ($[0.00,0.09]$) at target sparsity $0.5$
($p=0.015$, $d=1.3$). The robust result is the floor (Figure~\ref{fig:t1}): uniform quantisation
cannot reach low sparsity because the heavy-tailed integer weights round to zero, giving an
effective-sparsity floor of $0.73$ (int4) and $0.98$ (ternary), against $0.27$ and $0.32$ under
engram-aware mixed precision. These floors are deterministic properties of the weight
distribution. The structured advantage itself is regime-dependent and does not reproduce under an
independent sigmoid-conductance model, so we retain only the floor claim.

\section{Study 2: Replay-sequence compression}\label{sec:t3}
\begin{figure}[t]
\centering
\includegraphics[width=0.95\linewidth]{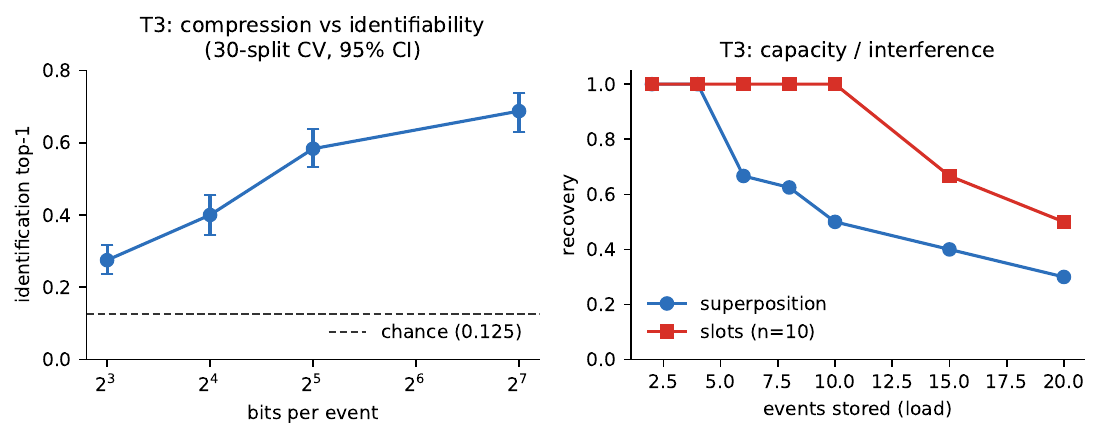}
\caption{Left: identification versus latent bit budget (30-split CV, 95\% CI; dashed line is
chance). Right: recovery from a fixed-size store versus load.}
\label{fig:t3}
\end{figure}

On hippocampal replay data (DANDI 000044 \citep{grosmark2016diversity,diba2007forward};
48 pyramidal units), place-ordered sequences carry decodable structure (Bayesian decode $0.34$
versus a time-shuffled null $0.20$). Under 30-split cross-validation, an uncompressed autoencoder
identifies held-out events at top-1 $0.69$ ($[0.63,0.74]$) against chance $0.125$ ($p=10^{-4}$),
degrading gradually with the bit budget (Figure~\ref{fig:t3}, left): a $48\times25$ sequence
($\sim38$~kbit) remains well above chance down to $8$--$16$ bits per event ($\sim10^3\times$
compression). A fixed-size incremental store (Figure~\ref{fig:t3}, right) shows a genuine capacity
limit, graceful for a superposition store and cliff-like for a slot store.

\section{Study 3: The migration pipeline in simulation}\label{sec:stages}
\begin{figure}[t]
\centering
\includegraphics[width=0.46\linewidth]{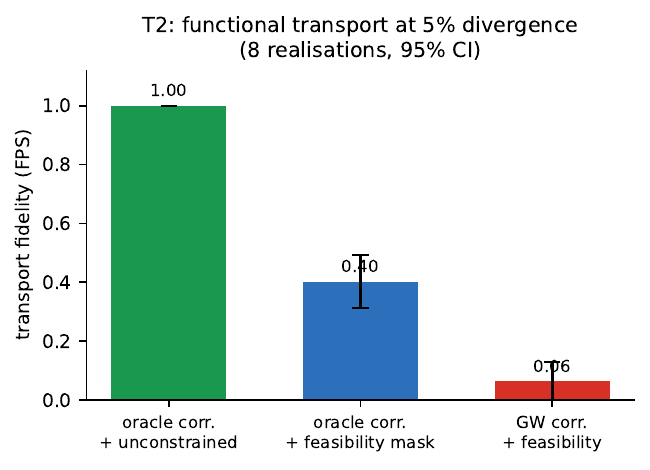}\hfill
\includegraphics[width=0.46\linewidth]{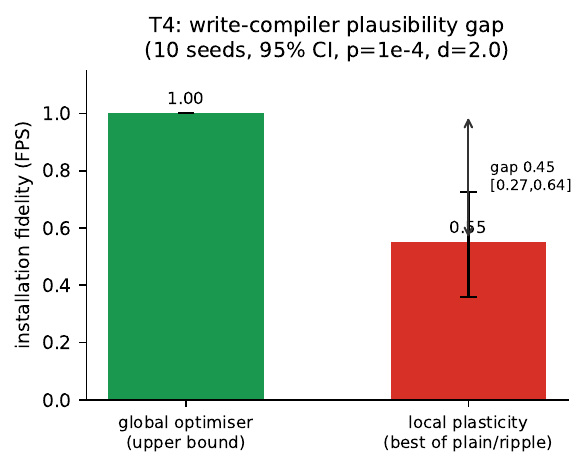}
\caption{Left: transport fidelity at $5\%$ divergence under three conditions (95\% CI). Right: the
write-compiler plausibility gap between a global optimiser and local plasticity.}
\label{fig:stages}
\end{figure}

Each stage tests a biological stage for which no direct evidence exists; we ask only whether it is
coherent in simulation. \textbf{Partition.} Holding the sensory/motor interface fixed, ablating
the asserted interneuron payload is no more disruptive than ablating a random interneuron subset
of the same size ($p=0.17,0.16,0.27,0.37$ across payload fractions $0.1$--$0.4$; 500-sample null):
there is no significantly separable identity module. \textbf{Transport.} Entropic
Gromov--Wasserstein recovers $0.76$ ($[0.74,0.78]$) of a ground-truth correspondence at $5\%$
divergence, but transport fidelity collapses from $1.00$ (oracle, unconstrained) to $0.40$
($[0.31,0.49]$, $p=3\times10^{-4}$) once the target's wiring is respected, and to $0.06$
($[0.00,0.13]$, $p=2\times10^{-4}$) under the estimated correspondence (Figure~\ref{fig:stages},
left). \textbf{Writing.} A global optimiser reaches ceiling $1.00$ on the target topology, while
strictly local reward-modulated plasticity reaches $0.55$ ($[0.36,0.73]$), a plausibility gap of
$0.45$ ($[0.27,0.64]$, $p=10^{-4}$, $d=2.0$), with ripple timing conferring no advantage
(Figure~\ref{fig:stages}, right). The stages fail for independent reasons; the target substrate's
wiring is the limit common to transport and writing.

\section{Discussion and limitations}
FPS cleanly separates the pipeline: compression of neural data is measurable and bounded, whereas
migration of function across substrates is unsupported by the partition--align--write pipeline,
with the target substrate's wiring as the recurring obstacle. Limitations are material: small
scale (a 306-neuron connectome, $\leq$135-neuron activity, 40-neuron networks); simple rate
dynamics; small test sets in places; a divergence model rather than a second organism in Study 2;
and FPS as a functional measure rather than a claim about subjective experience. The biological
prerequisites for the full pipeline are absent and not addressed.

\section{Reproducibility}\label{sec:repro}
All results are deterministic and run on a laptop. Each claim maps to a committed result artifact
and the script that produced it; figures are generated from those artifacts. Statistically
hardened artifacts (bootstrap CIs, permutation $p$-values) accompany each headline claim. Code,
data loaders, and the full operation set are released at
\url{https://github.com/Praveg432/functional-preservation-score}.

\bibliographystyle{tmlr}
\bibliography{references}

\appendix
\section{Biological claim provenance}\label{app:prov}
Torpor sheds $\sim$50--65\% of synapses with memory preserved \citep{vonderohe2006torpor},
supporting structured rather than uniform sparsification. Memory prosthetics
\citep{hampson2018prosthetic} restore recall within a subject, not across subjects. High-density
wireless neural interfaces \citep{jung2025bisc} record at the subdural surface, not synaptic
resolution. Whole-brain calcium imaging \citep{kato2015global} and the signal-propagation atlas
\citep{randi2023signal} provide the functional substrates used here. None of these supports the
non-destructive synaptic-resolution reading, identity-scale writing, or cross-subject transfer
the full pipeline would require.

\end{document}